\documentclass{article}

\usepackage{amsmath,amsfonts,bm}
\allowdisplaybreaks

\newcommand{\graph}{\mathcal{G}}
\newcommand{\LSTM}{\mathrm{LSTM}}

\newcommand{\MLP}{\mathrm{MLP}}
\newcommand{\MPN}{\mathrm{MPN}}
\newcommand{\attention}{\mathrm{attention}}

\newcommand{\set}[1]{\{ #1 \}}

\def\eqref#1{equation~\ref{#1}}
\def\1{\bm{1}}

\def\vmu{{\bm{\mu}}}
\def\vnu{{\bm{\nu}}}
\def\valpha{{\bm{\alpha}}}

\def\vdelta{{\bm{\delta}}}
\def\vsigma{{\bm{\sigma}}}
\def\vSigma{{\bm{\Sigma}}}

\def\vb{{\bm{b}}}
\def\vc{{\bm{c}}}

\def\ve{{\bm{e}}}
\def\vf{{\bm{f}}}

\def\vh{{\bm{h}}}
\def\vi{{\bm{i}}}

\def\vo{{\bm{o}}}
\def\vp{{\bm{p}}}
\def\vq{{\bm{q}}}

\def\vx{{\bm{x}}}
\def\vy{{\bm{y}}}
\def\vz{{\bm{z}}}

\def\mW{{\bm{W}}}

\DeclareMathAlphabet{\mathsfit}{\encodingdefault}{\sfdefault}{m}{sl}
\SetMathAlphabet{\mathsfit}{bold}{\encodingdefault}{\sfdefault}{bx}{n}

\def\gA{{\mathcal{A}}}

\def\gD{{\mathcal{D}}}
\def\gE{{\mathcal{E}}}
\def\gF{{\mathcal{F}}}
\def\gG{{\mathcal{G}}}
\def\gH{{\mathcal{H}}}

\def\gM{{\mathcal{M}}}
\def\gN{{\mathcal{N}}}

\def\gQ{{\mathcal{Q}}}

\def\gS{{\mathcal{S}}}

\def\gV{{\mathcal{V}}}

\newcommand{\softmax}{\mathrm{softmax}}
\newcommand{\sigmoid}{\sigma}

\usepackage{hyperref}       % hyperlinks
\usepackage{url}            % simple URL typesetting
\usepackage{booktabs}       % professional-quality tables
\usepackage{algorithm}
\usepackage{algorithmic}

\usepackage{amsfonts}       % blackboard math symbols
\usepackage{nicefrac}       % compact symbols for 1/2, etc.
\usepackage{microtype}      % microtypography
\usepackage{enumitem}
\usepackage{multirow}
\usepackage{graphicx}
\usepackage{cleveref}
\usepackage{subcaption}

\usepackage{hyperref}

\usepackage[accepted]{icml2020}

\icmltitlerunning{Hierarchical Generation of Molecular Graphs using Structural Motifs}

\begin{document}
\twocolumn[
\icmltitle{Hierarchical Generation of Molecular Graphs using Structural Motifs}

\begin{icmlauthorlist}
\icmlauthor{Wengong Jin}{mit}
\icmlauthor{Regina Barzilay}{mit}
\icmlauthor{Tommi Jaakkola}{mit}
\end{icmlauthorlist}

\icmlaffiliation{mit}{MIT CSAIL}

\icmlcorrespondingauthor{Wengong Jin}{wengong@csail.mit.edu}

\icmlkeywords{Machine Learning}

\vskip 0.3in
]


\printAffiliationsAndNotice{}  % leave blank if no need to mention equal contribution

\begin{abstract}
Graph generation techniques are increasingly being adopted for drug discovery. Previous graph generation approaches have utilized relatively small molecular building blocks such as atoms or simple cycles, limiting their effectiveness to smaller molecules. Indeed, as we demonstrate, their performance degrades significantly for larger molecules. In this paper, we propose a new hierarchical graph encoder-decoder that employs significantly larger and more flexible graph motifs as basic building blocks. Our encoder produces a multi-resolution representation for each molecule in a fine-to-coarse fashion, from atoms to connected motifs. Each level integrates the encoding of constituents below with the graph at that level. Our autoregressive coarse-to-fine decoder adds one motif at a time, interleaving the decision of selecting a new motif with the process of resolving its attachments to the emerging molecule. We evaluate our model on multiple molecule generation tasks, including polymers, and show that our model significantly outperforms previous state-of-the-art baselines.
\end{abstract}

\section{Introduction}

\begin{figure*}[t]
    \centering
    \includegraphics[width=\textwidth]{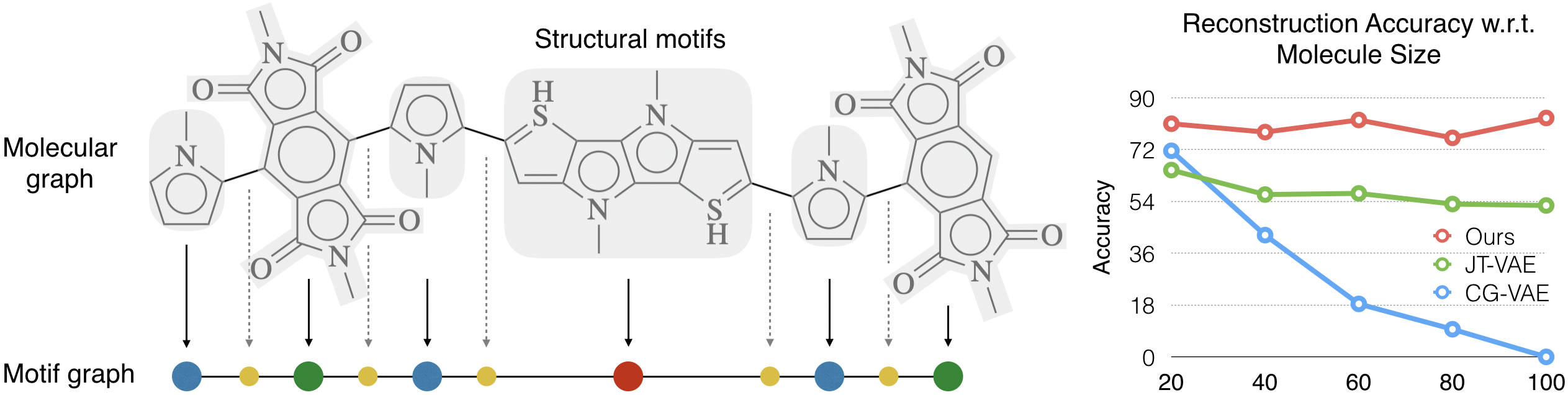}
    \caption{\textbf{Left}: Illustration of structural motifs in polymers. \textbf{Right}: Reconstruction accuracy for polymers with various sizes (number of atoms). Notably, the atom-based generative model CG-VAE~\citep{liu2018constrained} fails to reconstruct molecules over 80 atoms. In contrast, the proposed model maintains high accuracy for large molecules by utilizing motifs as building blocks for generation (red curve).}
    \label{fig:motivation}
    \vspace{-3pt}
\end{figure*}

%Drug discovery seeks to find novel compounds with desired biochemical properties. 
%This inverse design problem can be formulated as a graph generation task.
%Graph generation is computationally challenging due to discrete nature of graphs and complex dependencies involved in the joint distribution over nodes and edges.
%Similar to image and text generation, success in this task is predicated on the inductive biases built into the encoder-decoder architecture.

Deep learning models for molecule property prediction and molecule generation are improving at a fast pace. Work to date has adopted primarily two types of building blocks for representing and building molecules: atom-by-atom strategies~\citep{li2018learning,you2018graph,liu2018constrained}, or substructure based (either rings or bonds)~\citep{jin2018junction,jin2018learning}. While these methods have been successful for small molecules, their performance degrades significantly for larger molecules such as polymers (see Figure~\ref{fig:motivation}). The failure is likely due to many generation steps required to realize larger molecules and the associated challenges with gradients across the iterative steps. 

Large molecules such as polymers exhibit clear hierarchical structure, being built from repeated structural motifs. We hypothesize that explicitly incorporating such motifs as building blocks in the generation process can significantly improve reconstruction and generation accuracy, as already illustrated in Figure~\ref{fig:motivation}.
While different substructures as building blocks were considered in previous work~\citep{jin2018junction}, their approach could not scale to larger motifs. Indeed, their decoding process required each substructure neighborhood to be assembled in one go, making it combinatorially challenging to handle large components with many possible attachment points.

In this paper, we propose a motif-based hierarchical encoder-decoder for graph generation. The motifs themselves are extracted separately at the outset from frequently occurring substructures, regardless of size. During generation, molecules are built step by step by attaching motifs, large or small, to the emerging molecule. The decoder operates hierarchically, in a coarse-to-fine manner, and makes three key consecutive predictions in each pass: new motif selection, which part of it attaches, and the points of contact with the current molecule. These  decisions are highly coupled and naturally modeled auto-regressively. Moreover, each decision is directly guided by the information explicated in the associated layer of the mirroring hierarchical encoder. The feed-forward fine-to-coarse encoding performs iterative graph convolutions at each level, conditioned on the results from layer below. 

The proposed model is evaluated on various tasks ranging from polymer generative modeling to graph translation for molecule property optimization. Our baselines include state-of-the-art graph generation methods~\citep{you2018graph,liu2018constrained,jin2018learning}. 
On polymer generation, our model achieved state-of-the art results under various metrics, outperforming the best baselines with 20\% absolute improvement in reconstruction accuracy. On graph translation tasks, our model outperformed all the baselines, yielding 3.3\% and 8.1\% improvement on QED and DRD2 optimization tasks. 
During decoding, our model runs 6.3 times faster than previous substructure-based methods~\citep{jin2018learning}.
We further conduct ablation studies to validate the advantage of using larger motifs and model architecture.
\section{Background and Motivation}
\label{sec:motivation}

Molecules are represented as graphs $\gG = (\gV, \gE)$ with atoms $\gV$ as nodes and bonds $\gE$ as edges. Graphs are challenging objects to generate, especially for larger molecules such as polymers. For the polymer dataset used in our experiment, there are thousands of molecules with more than 80 atoms. 
To illustrate the challenge, we tested two state-of-the-art variational autoencoders~\citep{liu2018constrained,jin2018junction} on this dataset and found these models often fail to reconstruct molecules from their latent embedding (see Figure~\ref{fig:motivation}). 

The reason of this failure is that these methods generate molecules based on small building blocks. In terms of autoregressive models, previous work on molecular graph generation can be roughly divided in two categories:\footnote{We restrict our discussion to molecule generation. \citet{you2018graphrnn,liao2019efficient} developed generative models for other types of graphs such as social networks. Their current implementations do not support the prediction of node and edge attributes and cannot be directly applied to molecules. Thus their methods are not tested here.}
\begin{itemize}[leftmargin=*,topsep=0pt,itemsep=0pt]
    \item Atom-based methods~\citep{li2018learning,you2018graph,liu2018constrained} generate molecules atom by atom.
    \item Substructure-based methods~\citep{jin2018junction,jin2018learning} generates molecules based on small substructures restricted to rings and bonds (often no more than six atoms).
\end{itemize}
As the building blocks are typically small, it requires many decoding steps for current models to reconstruct polymers. Therefore they are prone to make errors when generating large molecules.
On the other hand, many of these molecules consist of structural motifs beyond simple substructures. The number of decoding steps can be significantly reduced if graphs are generated motif by motif. As shown in Figure~\ref{fig:motivation}, our motif-based method achieves a much higher reconstruction accuracy.

\textbf{Motivation for New Architecture }
Current substructure-based method~\citep{jin2018junction} requires a combinatorial enumeration to assemble substructures whose time complexity is exponential to substructure size. Their enumeration algorithm assumes the substructures to be of certain types (single cycles or bonds). In practice, their method often fails when handling rings with more than 10 atoms (e.g., memory error).
Unlike substructures, motifs are typically much larger and can have flexible structures (see Figure~\ref{fig:motivation}).
As a result, this method cannot be directly extended to utilize motifs in practice.

To this end, we propose a hierarchical encoder-decoder for graph generation. Our decoder allows arbitrary types of motifs and can assemble them efficiently without combinatorial explosion. Our encoder learns a hierarchical representation that allows the decoding process to depend on both coarse-grained motif and fine-grained atom connectivity.

\subsection{Motif Extraction}
We define a motif $\gS_i=(\gV_i,\gE_i)$ as a subgraph of molecule $\graph$ induced by atoms in $\gV_i$ and bonds in $\gE_i$. Given a molecule, we extract its motifs $\gS_1,\cdots,\gS_n$ such that their union covers the entire molecular graph: $\gV = \bigcup_i \gV_i$ and $\gE = \bigcup_i \gE_i$.
To extract motifs, we decompose a molecule $\graph$ into disconnected fragments by breaking all the bridge bonds that will not violate chemical validity (illustrations in the appendix).
\begin{enumerate}[leftmargin=*,topsep=0pt,itemsep=0pt]
    \item Find all the \emph{bridge} bonds $(u,v) \in \gE$, where both $u$ and $v$ have degree $\Delta_u, \Delta_v \geq 2$ and either $u$ or $v$ is part of a ring. Detach all the bridge bonds from its neighbors.
    \item Now the graph $\graph$ becomes a set of disconnected subgraphs $\graph_1, \cdots, \graph_N$. Select $\graph_i$ as motif in $\graph$ if its occurrence in the training set is more than $\Delta=100$. 
    \item If $\graph_i$ is not selected as motif, further decompose it into rings and bonds and select them as motif in $\graph$.
\end{enumerate}
We apply the above procedure to all the molecules in the training set and construct a vocabulary of motifs $V_\gS$. In the following section, we will describe how we encode and decode molecules using the extracted motifs.

\section{Hierarchical Graph Generation}

\begin{figure}[t]
    \centering
    \includegraphics[width=0.472\textwidth]{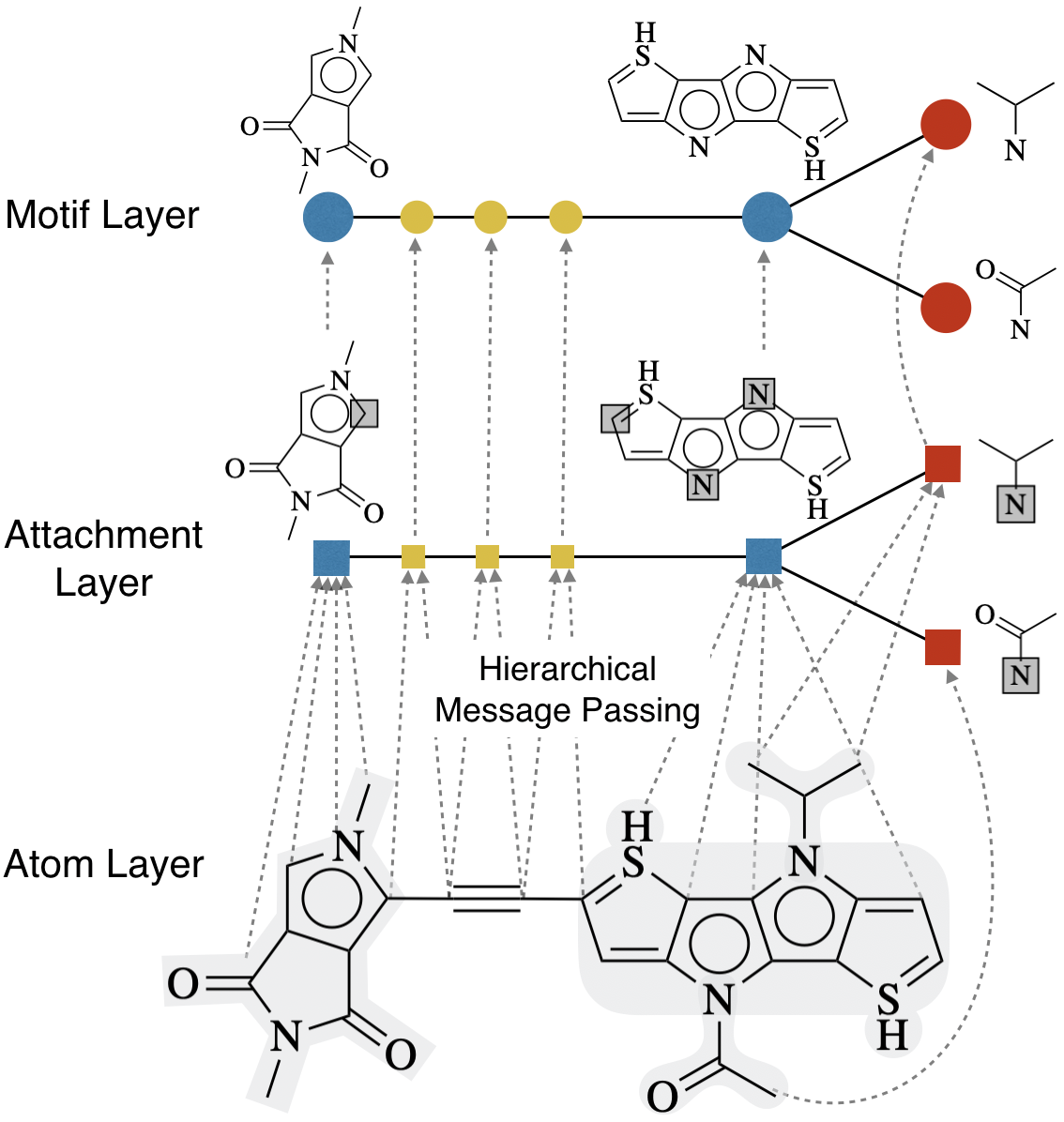}
    \caption{Hierarchical graph encoder. Dashed arrows connect each atom to the motifs it belongs. In the attachment layer, each node $\gA_i$ is a particular attachment configuration of motif $\gS_i$. The atoms in the intersection between each motif and its neighbors are highlighted in faded block.}
    \label{fig:encoder}
\end{figure}

\begin{figure}[t]
    \centering
    \includegraphics[width=0.47\textwidth]{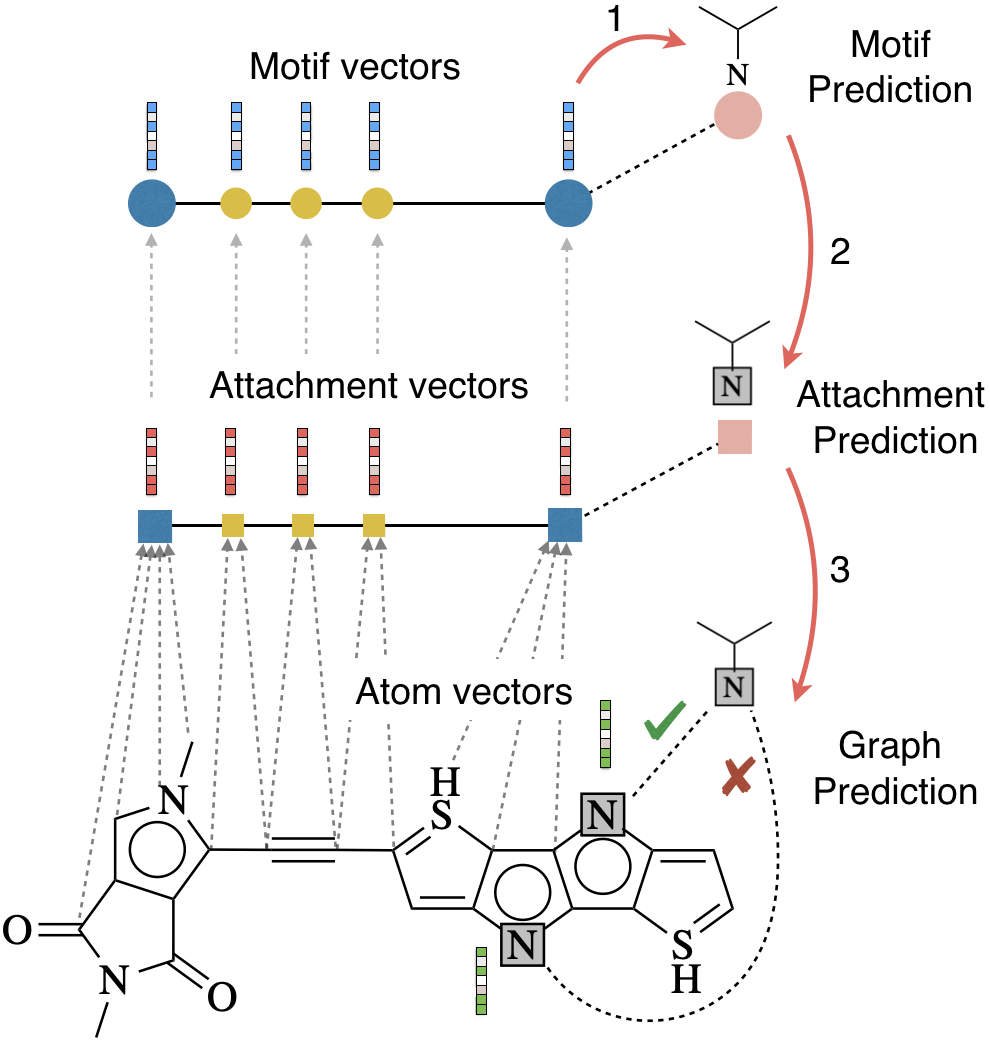}
    \caption{Hierarchical graph decoder. In each step, the decoder first runs hierarchical message passing to compute motif, attachment and atom vectors. Then it performs motif and attachment prediction for the next motif node. Finally, it decides how the new motif should be attached to the current graph via graph prediction.}
    \label{fig:decoder}
\end{figure}

Our approach extends the variational autoencoder~\citep{kingma2013auto} to molecular graphs by introducing a hierarchical decoder and a matching encoder.
In our framework, the probability of a graph $\graph$ is modeled as a joint distribution over structural motifs $\gS_1,\cdots,\gS_n$ constituting $\graph$, together with their attachments $\gA_1,\cdots,\gA_n$. Each attachment $\gA_i=\set{v_j \; | \; v_j \in \bigcup_k \gS_i \cap \gS_k}$ indicates the intersecting atoms between $\gS_i$ and its neighbor motifs.
To capture complex dependencies involved in the joint distribution of motifs and their attachments, we propose an auto-regressive factorization of $P(\graph)$:
\begin{equation}
    P(\graph) = \int_\vz P(\vz) \prod_k\nolimits P(\gS_k, \gA_k \;|\; \gS_{<k}, \gA_{<k}, \vz) d\vz
\end{equation} 
As illustrated in Figure~\ref{fig:decoder}, in each generation step, our decoder adds a new motif $\gS_k$ (\emph{motif prediction}) and its attachment configuration $\gA_k$ (\emph{attachment prediction}). Then it decides how the new motif should be attached to the current graph (\emph{graph prediction}).

To support the above hierarchical generation, we need to design a matching encoder representing molecules at multiple resolutions in order to provide necessary information for each decoding step. 
Therefore, we propose to represent a molecule $\graph$ by a hierarchical graph $\gH_\graph$ with three layers (see Figure~\ref{fig:encoder}): 
\begin{enumerate}[leftmargin=*,topsep=0pt,itemsep=0pt] 
    \item \textbf{Motif layer}: This layer represents how the motifs are coarsely connected in the graph. This layer provides essential information for the motif prediction in the decoding process. Specifically, this layer contains $n$ nodes $\gS_1,\cdots, \gS_n$ and $m$ edges $\{(\gS_i, \gS_j) \;|\; \gS_i \cap \gS_j \neq \emptyset\}$ for all intersecting motifs $\gS_i, \gS_j$. This layer is tree-structured due to our way of constructing motifs. 

    \item \textbf{Attachment layer}: This layer encodes the connectivity between motifs at a fine-grained level. Each node $\gA_i=(\gS_i,\set{v_j})$ in this layer represents a particular attachment configuration of motif $\gS_i$, where $\set{v_j}$ are atoms in the intersection between $\gS_i$ and one of its neighbor motifs (see Figure~\ref{fig:encoder}). This layer provides crucial information for the attachment prediction step during decoding, which helps reducing the space of candidate attachments between $\gS_i$ and its neighbor motifs.
    Just like the motif vocabulary $V_\gS$, all the attachment configurations of $\gS_i$ form a motif-specific vocabulary $V_\gA(\gS_i)$, which is computed from the training set.\footnote{In our experiments, the average size of attachment vocabulary $|V_\gA(\gS_i)| \leq 10$ and the size of motif vocabulary $|V_\gS|<500$.}
    
    \item \textbf{Atom layer}: The atom layer is the molecular graph $\graph$ representing how its atoms are connected. Each atom node $v$ is associated with a label $a_v$ indicating its atom type and charge. Each edge $(u,v)$ in the atom layer is labeled with $b_{uv}$ indicating its bond type. This layer provides necessary information for the graph prediction step during decoding.
\end{enumerate}

We further introduce edges that connect the atoms and motifs between different layers in order to propagate information in between. In particular, we draw a directed edge from atom $v$ in the atom layer to node $\gA_i$ in the attachment layer if $v \in \gS_i$. We also draw edges from $\gA_i$ to $\gS_i$ in the motif layer. This gives us the hierarchical graph $\gH_\graph$ for molecule $\graph$, which will be encoded by a hierarchical message passing network (MPN). During encoding, each node $\gS_i$ is represented as a one-hot encoding in the motif vocabulary $V_\gS$. Likewise, each node $\gA_i$ is represented as a one-hot encoding in the attachment vocabulary $V_\gA(\gS_i)$.

\subsection{Hierarchical Graph Encoder}
\label{sec:encoder}

Our encoder contains three MPNs that encode each of the three layers in the hierarchical graph. For simplicity, we denote the MPN encoding process as $\MPN_\psi(\cdot)$ with parameter $\psi$, and denote $\MLP(\vx, \vy)$ as a multi-layer neural network whose input is the concatenation of $\vx$ and $\vy$. The details of MPN architecture is listed in the appendix.

\textbf{Atom Layer MPN } 
We first encode the atom layer of $\gH_\graph$ (denoted as $\gH_\graph^g$). The inputs to this MPN are the embedding vectors $\set{\ve(a_u)}, \set{\ve(b_{uv})}$ of all the atoms and bonds in $\graph$. During encoding, the network propagates the message vectors between different atoms for $T$ iterations and then outputs the atom representation $\vh_v$ for each atom $v$:
\begin{equation}
\vc_\graph^g = \set{\vh_v} = \MPN_{\psi_1}\left(\gH_\graph^g, \set{\ve(a_u)}, \set{\ve(b_{uv})} \right)
\end{equation}

\textbf{Attachment Layer MPN } 
The input feature of each node $\gA_i$ in the attachment layer $\gH_\graph^a$ is an concatenation of the embedding $\ve(\gA_i)$ and the sum of its atom vectors $\set{\vh_v \;|\; v \in \gS_i}$: 
\begin{equation}
\vf_{\gA_i} = \MLP\left( \ve(\gA_i), \sum\nolimits_{v \in \gS_i} \vh_v \right) 
\end{equation}
The input feature for each edge $(\gA_i, \gA_j)$ in this layer is an embedding vector $\ve(d_{ij})$, where $d_{ij}$ describes the relative ordering between node $\gA_i$ and $\gA_j$ during decoding. Specifically, we set $d_{ij}=k$ if node $\gA_i$ is the $k$-th child of node $\gA_j$ and $d_{ij}=0$ if $\gA_i$ is the parent. We then run $T$ iterations of message passing over $\gH_\graph^a$ to compute the motif representations:
\begin{equation}
\vc_\graph^a = \set{\vh_{\gA_i}} = \MPN_{\psi_2}\left(\gH_\graph^a, \set{\vf_{\gA_i}}, \set{\ve(d_{ij})} \right)
\label{eq:mpn}
\end{equation}

\textbf{Motif Layer MPN } Similarly, the input feature of node $\gS_i$ in this layer is computed as the concatenation of embedding $\ve(\gS_i)$ and the node vector $\vh_{\gA_i}$ from the previous layer. Finally, we run message passing over the motif layer $\gH_\graph^s$ to obtain the motif representations:
\begin{eqnarray}
\vf_{\gS_i} &=& \MLP\left( \ve(\gS_i), \vh_{\gA_i} \right) \\
\vc_\graph^s = \set{\vh_{\gS_i}} &=& \MPN_{\psi_3}\left(\gH_\graph^s, \set{\vf_{\gS_i}}, \set{\ve(d_{ij})} \right)
\end{eqnarray}
Finally, we represent a molecule $\graph$ by a latent vector $\vz_\graph$ sampled through reparameterization trick with mean $\vmu(\vh_{\gS_1})$ and log variance $\vSigma(\vh_{\gS_1})$:
\begin{equation}
    \vz_\graph = \vmu(\vh_{\gS_1}) + \exp (\vSigma(\vh_{\gS_1})) \cdot \epsilon; \quad \epsilon \sim \gN(\mathbf{0}, \mathbf{I})
\end{equation}
where $\gS_1$ is the root motif (i.e., the first motif to be generated during reconstruction). %In section $\S_1$, we will discuss how to aggregate vectors $\vc_\graph$ through attention mechanism for graph-to-graph translation tasks. 

\subsection{Hierarchical Graph Decoder}
\label{sec:decoder}

As illustrated in Figure~\ref{fig:decoder}, our graph decoder generates a molecule $\graph$ by incrementally expanding its hierarchical graph. In $t^\mathrm{th}$ generation step, we first use the same hierarchical MPN architecture to encode all the motifs and atoms in $\gH_{\graph}^{(t)}$, the (partial) hierarchical graph generated till step $t$. This gives us motif vectors $\vh_{\gS_k}$ and atom vectors $\vh_{v_j}$ for the existing motifs and atoms.

During decoding, the model maintains a set of frontier nodes $\gF$ where each node $\gS_k \in \gF$ is a motif that still has neighbors to be generated. $\gF$ is implemented as a stack because motifs are generated in their depth-first order.
Suppose $\gS_k$ is at the top of stack $\gF$ in step $t$, the model makes the following predictions conditioned on latent representation $\vz_\graph$:
\begin{enumerate}[leftmargin=*,topsep=0pt,itemsep=0pt]
    \item \textbf{Motif Prediction}: The model predicts the next motif $\gS_t$ to be attached to $\gS_k$. This is cast as a classification task over the motif vocabulary $V_\gS$:
    \begin{equation}
        \vp_{\gS_t} = \softmax(\MLP(\vh_{\gS_k}, \vz_\graph)) \label{eq:motif-predict}
    \end{equation}
    
    \item \textbf{Attachment Prediction}: Now the model needs to predict the attachment configuration $\gA_t$ of motif $\gS_t$ (i.e., what atoms $v_j \in \gS_t$ belong to the intersection of $\gS_t$ and its neighbor motifs). This is also cast as a classification task over the attachment vocabulary $V_\gA(\gS_t)$:
    \begin{equation}
        \vp_{\gA_t} = \softmax(\MLP(\vh_{\gS_k}, \vz_\graph)) \label{eq:attach-predict}
    \end{equation}
    This prediction step is crucial because it significantly reduces the space of possible attachments between $\gS_t$ and its neighbor motifs.
    
    \item \textbf{Graph Prediction}: Finally, the model must decide how $\gS_t$ should be attached to $\gS_k$. The attachment between $\gS_t$ and $\gS_k$ is defined as atom pairs $\gM_{tk} = \set{(u_j,v_j) \;|\; u_j \in \gA_k, v_j \in \gA_t}$ where atom $u_j$ and $v_j$ are attached together. The probability of a candidate attachment $M$ is computed based on the atom vectors $\vh_{u_j}$ and $\vh_{v_j}$:
    \begin{eqnarray}
        \vp_M &=& \softmax\left(\vh_M \cdot \vz_\graph \right) \label{eq:attach-predict2} \\
        \vh_M &=& \sum_j\nolimits \MLP(\vh_{u_j},\vh_{v_j})
    \end{eqnarray}
    The number of possible attachments are limited because the number of attaching atoms between two motifs is small and the attaching points must be consecutive.\footnote{In our experiments, the number of possible attachments are usually less than 20 for polymers and small molecules.}
\end{enumerate}
The above three predictions together give an autoregressive factorization of the distribution over the next motif and its attachment. 
Each of the three decoding steps depends on the outcome of previous step, and predicted attachments will in turn affect the prediction of subsequent motifs.

\textbf{Training } During training, we apply teacher forcing to the above generation process, where the generation order is determined by a depth-first traversal over the ground truth molecule. Given a training set of molecules, we seek to minimize the negative ELBO:
\begin{equation}
    - \mathbb{E}_{\vz \sim Q}[\log P(\graph | \vz)] + \lambda_{\text{KL}}\gD_{\text{KL}}[Q(\vz| \graph) || P(\vz)] \label{eq:vae}
\end{equation}

\subsection{Extension to Graph-to-Graph Translation}
\label{sec:translation}
The proposed architecture can be naturally extended to graph-to-graph translation~\citep{jin2018learning} for molecular optimization, which seeks to modify compounds in order to improve their biochemical properties. 
Given a corpus of molecular pairs $\set{(X,Y)}$, where $Y$ is a structural analog of $X$ with better chemical properties, the model is trained to translate an input molecular graph into its better form. 
In this case, we seek to learn a translation model $P(Y | X)$ parameterized by our encoder-decoder architecture. We also introduce attention layers into our model, which is crucial for translation performance~\citep{bahdanau2014neural}.

\textbf{Training } In graph translation, a compound $X$ can be associated with multiple outputs $Y$ since there are many ways to modify $X$ to improve its properties. In order to generate diverse outputs, we follow previous work~\citep{zhu2017toward,jin2018learning} and incorporate latent variables $\vz$ to the translation model:
\begin{equation}
    P(Y | X) = \int_\vz P(Y | X, \vz) P(\vz) d\vz
\end{equation} 
where the latent vector $\vz$ indicates the intended mode of translation, sampled from a prior $P(\vz)$ during testing.

The model is trained as a conditional variational autoencoder. Given a training example $(X,Y)$, we sample $\vz$ from the approximate posterior $Q(\vz | X,Y) = \gN(\vmu_{X,Y}, \vsigma_{X,Y})$. To compute $Q(\vz | X,Y)$, we first encode $X$ and $Y$ into their representations $\vc_X$ and $\vc_Y$ and then compute difference vector $\vdelta_{X,Y}$ that summarizes the structural changes from molecule $X$ to $Y$ at both atom and motif level:
\begin{equation}
    \vdelta_{X,Y}^s = \sum \vc_Y^s - \sum \vc_X^s \quad 
    \vdelta_{X,Y}^g = \sum \vc_Y^g - \sum \vc_X^g \nonumber
    \label{eq:vae_diff}
\end{equation}
Finally, we compute $[\vmu_{X,Y}, \vsigma_{X,Y}] = \MLP(\vdelta_{X,Y}^\gS, \vdelta_{X,Y}^\graph)$ and sample $\vz$ using reparameterization trick. The latent code $\vz$ is passed to the decoder along with the input representation $\vc_X$ to reconstruct output $Y$. The training objective is to minimize negative ELBO similar to Eq.(\ref{eq:vae}).

\textbf{Attention } For graph translation, the input molecule $X$ is embedded by our hierarchical encoder into a set of vectors $\vc_X=\vc_X^s \cup \vc_X^a \cup \vc_X^g$, representing the molecule at multiple resolutions. These vectors are fed into the decoder through attention mechanisms~\citep{luong2015effective}. Specifically, we modify the motif prediction (Eq.~\ref{eq:motif-predict}) into
\begin{eqnarray}
    \vp_{\gS_t} &=& \softmax(\MLP(\vh_{\gS_k}, \valpha_k^s, \vz)) \\
    \valpha_k^s &=& \attention\left( \vh_{\gS_k}, \vc_X^s \right)
\end{eqnarray}
where $\attention(\vh_*, \vc_X^s)$ is a bilinear attention over vectors $\vc_X^s$ with query vector $\vh_{\gS_k}$. The attachment prediction (Eq.~\ref{eq:attach-predict}) is modified similarly with its attention over $\vc_X^a$. The graph prediction (Eq.~\ref{eq:attach-predict2}) is modified into
\begin{eqnarray}
    \vp_M &=& \softmax\left(\vh_M \cdot \attention(\vh_M, \vc_X^g)\right) \\
    \vh_M &=& \sum_j\nolimits \MLP(\vh_{u_j},\vh_{v_j}, \vz)
\end{eqnarray}

\section{Experiments}
\label{sec:experiment}
\newcommand\Tstrut{\rule{0pt}{2.3ex}}
\newcommand\Bstrut{\rule[-0.9ex]{0pt}{0pt}}

We evaluate our method on two application tasks. The first task is polymer generative modeling. This experiment validates our argument in section \ref{sec:motivation} that our model is advantageous when the molecules have large sizes. The second task is graph-to-graph translation for small molecules. Here we show the proposed architecture also brings benefits to small molecules compared to previous state-of-the-art graph generation methods. 

\subsection{Polymer Generative Modeling}

\begin{table*}[t]
\centering
\vspace{-3pt}
\caption{Results on polymer generative modeling. The first row reports the oracle performance using real data as generated samples. The last row (small motif) is a variant of our model where we restrict the motif vocabulary to contain only single rings and bonds (similar to JT-VAE). ``Recon.'' means reconstruction accuracy; ``Div.'' means diversity; SNN means nearest neighbor similarity; ``Frag / Scaf'' means fragment and scaffold similarity. Except property statistics, all metrics are the higher the better.}
\vspace{5pt}
\begin{tabular}{lccccccccccc}
\hline
\multirow{2}{*}{Method} & \multicolumn{4}{c}{ Reconstruction / Sample Quality ($\uparrow$) } & \multicolumn{4}{c}{Property Statistics ($\downarrow$)} & \multicolumn{3}{c}{Structural Statistics ($\uparrow$)} \Tstrut\Bstrut \\
\cline{2-12}
 & Recon. & Valid & Unique & Div. & logP & SA & QED & MW & SNN & Frag. & Scaf. \Tstrut\Bstrut \\
\hline
Real data & - & 100\% & 100\% & 0.823 & 0.094 & 6.7e-5 & 1.7e-5 & 82.3 & 0.706 & 0.995 & 0.462 \Tstrut\Bstrut \\
\hline
SMILES & 21.5\% & 93.1\% & \textbf{97.3\%} & 0.821 & 1.471 & 0.011 & \textbf{5.4e-4} & 4963 & 0.704 & 0.981 & 0.385 \Tstrut\Bstrut \\
CG-VAE & 42.4\% & \textbf{100\%} & 96.2\% & \textbf{0.879} & 3.958 & 2.600 & 0.0030 & 3944 & 0.204 & 0.372 & 0.001 \Tstrut\Bstrut \\
JT-VAE & 58.5\% & \textbf{100\%} & 94.1\% & 0.864 & 2.645 & 0.157 & 0.0075 & 10867 & 0.522 & 0.925 & 0.297 \Tstrut\Bstrut \\
\hline
HierVAE & \textbf{79.9\%} & \textbf{100\%} & 97.0\% & 0.817 & \textbf{0.525} & \textbf{0.007} & 5.7e-4 & \textbf{1928} & \textbf{0.708} & \textbf{0.984} & \textbf{0.390} \Tstrut\Bstrut \\
$\boldsymbol{\cdot}$ Small motif & 71.0\% & \textbf{100\%} & 97.2\% & 0.835 & 0.872 & 0.042 & 0.0019 & 5320 & 0.575 & 0.949 & 0.191 \Tstrut\Bstrut \\
\hline
\end{tabular}
\label{tab:polymer}
\end{table*}

\begin{figure}[t]
    \centering
    \includegraphics[width=0.48\textwidth]{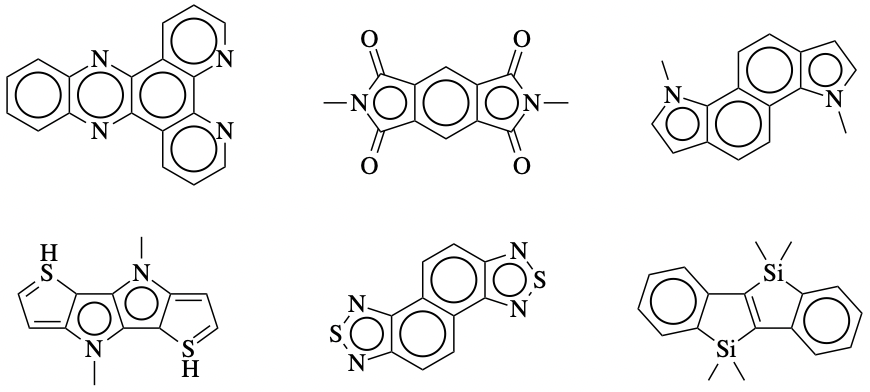}
    \caption{Examples of motif structures utilized by our model. These motifs consist of multiple rings and bonds, which are substantially more complex than previous methods~\citep{jin2018junction}.}
    \vspace{-2pt}
    \label{fig:motif}
\end{figure}

\textbf{Dataset } Our method is evaluated on the polymer dataset from \citet{st2019message}, which contains 86K polymers in total (after removing duplicates). The dataset is divided into 76K, 5K and 5K for training, validation and testing. Using our motif extraction, we collected 436 different motifs (examples shown in Figure~\ref{fig:motif}). On average, each motif has 5.24 different attachment configurations. The distribution of motif size and their frequencies are reported in Figure~\ref{fig:speed}.

\textbf{Evaluation Metrics } Our evaluation effort measures various aspects of molecule generation proposed in \citet{kusner2017grammar,polykovskiy2018molecular}. Besides basic metrics like chemical validity and diversity, we compare distributional statistics between generated and real compounds. A good generative model should generate molecules which present similar aggregate statistics to real compounds. Our metrics include (with details shown in the appendix):
\begin{itemize}[leftmargin=*,topsep=0pt,itemsep=0pt] 
    \item \textbf{Reconstruction accuracy}: We measure how often the model can completely reconstruct a given molecule from its latent embedding $\vz$. The reconstruction accuracy is computed over 5K compounds in the test set.
    \item \textbf{Validity}: Percentage of chemically valid compounds.
    \item \textbf{Uniqueness}: Percentage of unique compounds.
    \item \textbf{Diversity}: We compute the pairwise molecular distance among generated compounds. The molecular distance $\mathrm{dist}(X,Y)$ is defined as the Tanimoto distance over Morgan fingerprints~\citep{rogers2010extended} of two molecules.
    \item \textbf{Property statistics}: We compare property statistics between generated molecules and real data. Our properties include \emph{partition coefficient} (logP), \emph{synthetic accessibility} (SA), \emph{drug-likeness} (QED) and \emph{molecular weight} (MW). To quantitatively evaluate the distance between two distributions, we compute Frechet distance between property distributions of molecules in the generated and test sets~\citep{polykovskiy2018molecular}.
    \item \textbf{Structural statistics}: We also compute structural statistics between generated molecules and real data. \emph{Nearest neighbor similarity} (SNN) is the average similarity of generated molecules to the nearest molecule from the test set. \emph{Fragment similarity} (Frag) and \emph{scaffold similarity} (Scaf) are cosine distances between vectors of fragment or scaffold frequencies of the generated and the test set.
\end{itemize}

\textbf{Baselines } We compare our method against three state-of-the-art variational autoencoders for molecular graphs.
SMILES VAE~\citep{gomez2016automatic} is a sequence to sequence VAE that generates molecules based on their SMILES strings~\citep{weininger1988smiles}.
CG-VAE~\citep{liu2018constrained} is a graph-based VAE generating molecules atom by atom.
JT-VAE~\citep{jin2018junction} is also a graph-based VAE generating molecules based on simple substructures restricted to rings and bonds. 
Finally, we report the oracle performance of distributional statistics by using real molecules in the training set as our generated samples. 

\subsubsection{Results}

\begin{figure}[t]
    \centering
    \includegraphics[width=0.48\textwidth]{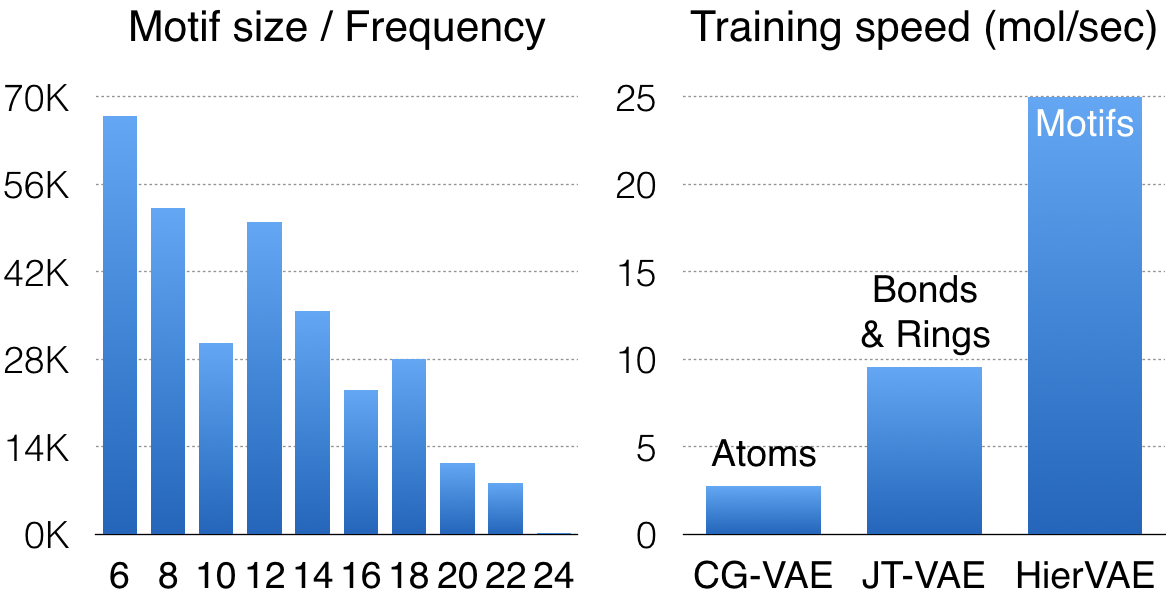}
    \caption{\textbf{Left}: Histogram of motif frequencies with respect to their sizes (i.e., number of atoms). \textbf{Right}: Training speed comparison between our method and baselines (on the same hardware).}
    \vspace{-2pt}
    \label{fig:speed}
    \vspace{-4pt}
\end{figure}

The performance of different methods are summarized in Table~\ref{tab:polymer}, Our method (HierVAE) significantly outperforms all previous methods in terms of reconstruction accuracy (79.9\% vs 58.5\%). This validates the advantage of utilizing large structural motifs, which reduces the number of generation steps. 
In terms of distributional statistics, our method achieves state-of-the-art results on logP (0.525 vs 1.471), molecular weight Frechet distance (1928 vs 4863) and all the structural similarity metrics.
Since our model requires fewer generation steps, our training speed is much faster than other graph-based methods (see Figure~\ref{fig:speed}).

\textbf{Ablation Study } To validate the importance of utilizing large structural motifs, we further experiment a variant of our model ($\mathrm{small\; motif}$), which keeps the same architecture but replaces the large structural motifs with basic substructures such as rings and bonds (with less than ten atoms). As shown in Table~\ref{tab:polymer}, its performance is significantly worse than our full model even though it builds on the same hierarchical architecture.

\subsection{Graph-to-Graph Translation}

\begin{table*}[t]
\centering
\vspace{-5pt}
\caption{Results on graph translation tasks from \citet{jin2018learning}. We report average improvement for continuous properties (logP), and success rate for binary properties (e.g., DRD2).}
\vspace{5pt}
\begin{tabular}{lcccccccc}
\hline
\multirow{2}{*}{Method} & \multicolumn{2}{c}{ logP ($\mathrm{sim} \geq 0.6$) } & \multicolumn{2}{c}{ logP ($\mathrm{sim} \geq 0.4$) } & \multicolumn{2}{c}{ Drug likeness } & \multicolumn{2}{c}{ DRD2 } \Tstrut\Bstrut \\
\cline{2-9}
& Improvement & Diversity & Improvement & Diversity & Success & Diversity & Success & Diversity \Tstrut\Bstrut \\
\hline
JT-VAE & $0.28 \pm 0.79$ & - & $1.03 \pm 1.39$ & - & 8.8\% & - & 3.4\% & - \Tstrut\Bstrut \\
CG-VAE & $0.25 \pm 0.74$ & - & $0.61 \pm 1.09$ & - & 4.8\% & - & 2.3\% & - \Tstrut\Bstrut \\
GCPN & $0.79 \pm 0.63$ & - & $2.49 \pm 1.30$ & - & 9.4\% & 0.216 & 4.4\% & 0.152 \Tstrut\Bstrut \\
MMPA & $1.65 \pm 1.44$ & 0.329 & $3.29 \pm 1.12$ & 0.496 & 32.9\% & 0.236 & 46.4\% & \textbf{0.275} \Tstrut\Bstrut \\
Seq2Seq & $2.33 \pm 1.17$ & 0.331 & $3.37 \pm 1.75$ & 0.471 & 58.5\% & 0.331 & 75.9\% & 0.176 \Tstrut\Bstrut \\
JTNN & $2.33 \pm 1.24$ & 0.333 & $3.55 \pm 1.67$ & 0.480 & 59.9\% & 0.373 & 77.8\% & 0.156 \Tstrut\Bstrut \\
AtomG2G & $2.41 \pm 1.19$ & 0.379 & \textbf{3.98 $\pm$ 1.54} & 0.563 & 73.6\% & 0.421 & 75.8\% & 0.128 \Tstrut\Bstrut \\
\hline
HierG2G & \textbf{2.49 $\pm$ 1.09} & \textbf{0.381} & \textbf{3.98 $\pm$ 1.46} & \textbf{0.564} & \textbf{76.9\%} & \textbf{0.477} & \textbf{85.9\%} & 0.192 \Tstrut\Bstrut \\
\hline
\end{tabular}
\label{tab:translation}
\vspace{-3pt}
\end{table*}

\begin{table}[t]
\centering
\vspace{-5pt}
\caption{Ablation study: the importance of hierarchical graph encoding, LSTM MPN architecture and structure-based decoding.}
\vspace{5pt}
\begin{tabular}{lcc}
\hline
Method & QED & DRD2  \Tstrut\Bstrut \\
\hline
HierG2G & \textbf{76.9\%} & \textbf{85.9\%}   \Tstrut\Bstrut \\
$\boldsymbol{\cdot}$ atom-based decoder & 76.1\% & 75.0\% \Tstrut\Bstrut \\
$\boldsymbol{\cdot}$ two-layer encoder & 75.8\% & 83.5\%  \Tstrut\Bstrut \\
$\boldsymbol{\cdot}$ one-layer encoder & 67.8\% & 74.1\% \Tstrut\Bstrut \\
%$\boldsymbol{\cdot}$ \small{GRU MPN}  & 72.6\% & 83.7\%  \Tstrut\Bstrut \\
\hline
\end{tabular}
\label{tab:ablation}
\vspace{-6pt}
\end{table}

We follow the experimental design by \citet{jin2018learning} and evaluate our  model on their graph-to-graph translation tasks.
Following their setup, we require the molecular similarity between $X$ and output $Y$ to be above certain threshold $\mathrm{sim}(X,Y) \geq \delta$ at test time. This is to prevent the model from ignoring input $X$ and translating it into arbitrary compound. Here the molecular similarity is defined as $\mathrm{sim}(X,Y)=1-\mathrm{dist}(X,Y)$.

\textbf{Dataset } The dataset consists of four property optimization tasks. In each task, we train and evaluate our model on their provided training and test sets.

\begin{itemize}[leftmargin=*,topsep=0pt,itemsep=0pt] 
\item \textbf{LogP}: The penalized logP score~\citep{kusner2017grammar} measures the solubility and synthetic accessibility of a compound. In this task, the model needs to translate input $X$ into output $Y$ such that $\mathrm{logP}(Y) > \mathrm{logP}(X)$. We experiment with two similarity thresholds $\delta=\set{0.4, 0.6}$.

\item \textbf{QED}: The QED score~\citep{bickerton2012quantifying} quantifies a compound's drug-likeness. In this task, the model is required to translate molecules with QED scores from the lower range $[0.7, 0.8]$ into the higher range $[0.9,1.0]$. The similarity constraint is $\mathrm{sim}(X,Y) \geq 0.4$.

\item \textbf{DRD2}: This task involves the optimization of a compound's biological activity against dopamine type 2 receptor (DRD2). The model needs to translate inactive compounds ($p < 0.05$) into active compounds ($p \geq 0.5$), where the bioactivity is assessed by a property prediction model from \citet{olivecrona2017molecular}. The similarity constraint is $\mathrm{sim}(X,Y) \geq 0.4$.
\end{itemize}

\textbf{Evaluation Metrics } Our evaluation metrics include translation accuracy and diversity. Each test molecule $X_i$ is translated $K=20$ times with different latent codes sampled from the prior distribution. On the logP optimization, we select compound $Y_i$ as the final translation of $X_i$ that gives the highest property improvement and satisfies $\mathrm{sim}(X_i,Y_i)\geq\delta$. We then report the average property improvement $\frac{1}{\gD}\sum_i\mathrm{logP}(Y_i)-\mathrm{logP}(X_i)$ over test set $\gD$. For other tasks, we report the translation success rate. A compound is successfully translated if one of its $K$ translation candidates satisfies all the similarity and property constraints of the task.
To measure the diversity, for each molecule we compute the average pairwise Tanimoto distance between all its successfully translated compounds.

\textbf{Baselines } We compare our method against the baselines including GCPN \citep{you2018graph}, MMPA \citep{dalke2018mmpdb} and translation based methods Seq2Seq and JTNN \citep{jin2018learning}.
Seq2Seq is a sequence-to-sequence model that generates molecules by their SMILES strings. 
JTNN is a graph-to-graph architecture that generates molecules structure by structure, but its decoder is not fully autoregressive. 

To make a direct comparison possible between our method and atom-based generation, we further developed an atom-based translation model (AtomG2G) as baseline. 
It makes three predictions in each generation step. First, it predicts whether the decoding process has completed (no more new atoms). If not, it creates a new atom $a_t$ and predicts its atom type. Lastly, it predicts the bond type between $a_t$ and other atoms autoregressively to fully capture edge dependencies~\citep{you2018graphrnn}.
The encoder of AtomG2G encodes only the atom-layer graph and the decoder attention only sees the atom vectors $\vc_X^\graph$. All translation models are trained under the same variational objective. Details of baseline architectures are in the appendix. 

\subsubsection{Results}
As shown in Table~\ref{tab:translation}, our model (HierG2G) achieves the new state-of-the-art on the four translation tasks. In particular, our model significantly outperforms JTNN in both translation accuracy (e.g., 76.9\% versus 59.9\% on the QED task) and output diversity (e.g., 0.564 versus 0.480 on the logP task). While both methods generate molecules by structures, our decoder is autoregressive which can learn more expressive mappings. 
In addition, our model runs 6.3 times faster than JTNN during decoding. 
Our model also outperforms AtomG2G on three datasets, with over 10\% improvement on the DRD2 task. This shows the advantage of our hierarchical model.

\textbf{Ablation Study } To understand the importance of different architecture choices, we report ablation studies over the QED and DRD2 tasks in Table~\ref{tab:ablation}. We first replace our hierarchical decoder with the atom-based decoder of AtomG2G to see how much the motif-based decoding benefits us. We keep the same hierarchical encoder but modified the input of the decoder attention to include both atom and motif vectors. Using this setup, the model performance decreases by 0.8\% and 10.9\% on the two tasks. We suspect the DRD2 task benefits more from motif-based decoding because biological target binding often depends on the presence of specific functional groups.

Our second experiment reduces the number of hierarchies in our encoder and decoder MPN, while keeping the same hierarchical decoding process. When the top motif layer is removed, the translation accuracy drops slightly by 0.8\% and 2.4\%. When we further remove the attachment layer (\emph{one-layer encoder}), the performance degrades significantly on both datasets. This is because all the motif information is lost and the model needs to infer what motifs are and how motif layers are constructed for each molecule.
This shows the importance of the hierarchical representation.
%Lastly, we replaced our LSTM MPN with the original GRU MPN used in JTNN. While the translation performance decreased by 4\% and 2.2\%, our method still outperforms JTNN by a wide margin. Therefore we use the LSTM MPN architecture for both HierG2G and AtomG2G baseline. 

\section{Related Work}

\textbf{Graph Generation }
Previous work have adopted various approaches for generating molecular graphs.
\citet{gomez2016automatic,segler2017generating,kusner2017grammar,dai2018syntax-directed,guimaraes2017objective,olivecrona2017molecular,popova2018deep,kang2018conditional} generated molecules based on their SMILES strings \citep{weininger1988smiles}.
\citet{simonovsky2018graphvae,de2018molgan,ma2018constrained} developed generative models which output the adjacency matrices and node labels of the graphs at once. 
\citet{you2018graphrnn,li2018learning,samanta2018nevae,liu2018constrained,zhou2018optimization} proposed generative models which decode molecules sequentially node by node.
\citet{seff2019discrete} developed a edit-based model which generates molecules based on insertions and deletions.

Our model is closely related to \citet{liao2019efficient} which generate graphs one block of nodes and edges at a time. While their encoder operates on original graphs, our encoder operates on multiple hierarchies and learns multi-resolution representations of input graphs.
Our work is also closely related to \citet{jin2018junction,jin2018learning} that generate molecules based on substructures. Their decoder first generates a junction tree with substructures as nodes, and then predicts how the substructures should be attached to each other. Their substructure attachment process involves combinatorial enumeration and therefore their model cannot scale to substructures more complex than simple rings and bonds. In contrast, our model allows the motif to have flexible structures. 

\textbf{Graph Encoders } Graph neural networks have been extensively studied for graph encoding~\citep{scarselli2009graph,bruna2013spectral,li2015gated,niepert2016learning,kipf2016semi,hamilton2017inductive,lei2017deriving,velickovic2017graph,xu2018powerful}.
Our method is related to graph encoders for molecules \citep{duvenaud2015convolutional,kearnes2016molecular,dai2016discriminative,gilmer2017neural,schutt2017schnet}. Different to these approaches, our method represents molecules as hierarchical graphs spanning from atom-level to motif-level graphs.

Our work is most closely related to \citep{defferrard2016convolutional,ying2018hierarchical,gao2019graph} that learn to represent graphs in a hierarchical manner. In particular, \citet{defferrard2016convolutional} utilized graph coarsening algorithms to construct multiple layers of graph hierarchy and
\citet{ying2018hierarchical,gao2019graph} proposed to learn the graph hierarchy jointly with the encoding process.
Despite some differences, all of these methods learns the hierarchy for regression or classification tasks. In contrast, our hierarchy is constructed for efficient graph generation.% and a molecule can be encoded into multiple sets of vectors, each representing the input at different resolutions.
%Our graph encoder learns hierarchical representation of molecules but our focus is on graph generation. 
\section{Conclusion}
In this paper, we developed a hierarchical encoder-decoder architecture generating molecular graphs using structural motifs as building blocks. The experimental results show our model outperforms prior atom and substructure based methods in both small molecule and polymer domains.


\bibliography{main}

\begin{thebibliography}{48}
\providecommand{\natexlab}[1]{#1}
\providecommand{\url}[1]{\texttt{#1}}
\expandafter\ifx\csname urlstyle\endcsname\relax
  \providecommand{\doi}[1]{doi: #1}\else
  \providecommand{\doi}{doi: \begingroup \urlstyle{rm}\Url}\fi

\bibitem[Bahdanau et~al.(2014)Bahdanau, Cho, and Bengio]{bahdanau2014neural}
Bahdanau, D., Cho, K., and Bengio, Y.
\newblock Neural machine translation by jointly learning to align and
  translate.
\newblock \emph{arXiv preprint arXiv:1409.0473}, 2014.

\bibitem[Bickerton et~al.(2012)Bickerton, Paolini, Besnard, Muresan, and
  Hopkins]{bickerton2012quantifying}
Bickerton, G.~R., Paolini, G.~V., Besnard, J., Muresan, S., and Hopkins, A.~L.
\newblock Quantifying the chemical beauty of drugs.
\newblock \emph{Nature chemistry}, 4\penalty0 (2):\penalty0 90, 2012.

\bibitem[Bruna et~al.(2013)Bruna, Zaremba, Szlam, and LeCun]{bruna2013spectral}
Bruna, J., Zaremba, W., Szlam, A., and LeCun, Y.
\newblock Spectral networks and locally connected networks on graphs.
\newblock \emph{arXiv preprint arXiv:1312.6203}, 2013.

\bibitem[Dai et~al.(2016)Dai, Dai, and Song]{dai2016discriminative}
Dai, H., Dai, B., and Song, L.
\newblock Discriminative embeddings of latent variable models for structured
  data.
\newblock In \emph{International Conference on Machine Learning}, pp.\
  2702--2711, 2016.

\bibitem[Dai et~al.(2018)Dai, Tian, Dai, Skiena, and
  Song]{dai2018syntax-directed}
Dai, H., Tian, Y., Dai, B., Skiena, S., and Song, L.
\newblock Syntax-directed variational autoencoder for structured data.
\newblock \emph{arXiv preprint arXiv:1802.08786}, 2018.

\bibitem[Dalke et~al.(2018)Dalke, Hert, and Kramer]{dalke2018mmpdb}
Dalke, A., Hert, J., and Kramer, C.
\newblock mmpdb: An open-source matched molecular pair platform for large
  multiproperty data sets.
\newblock \emph{Journal of chemical information and modeling}, 2018.

\bibitem[De~Cao \& Kipf(2018)De~Cao and Kipf]{de2018molgan}
De~Cao, N. and Kipf, T.
\newblock Molgan: An implicit generative model for small molecular graphs.
\newblock \emph{arXiv preprint arXiv:1805.11973}, 2018.

\bibitem[Defferrard et~al.(2016)Defferrard, Bresson, and
  Vandergheynst]{defferrard2016convolutional}
Defferrard, M., Bresson, X., and Vandergheynst, P.
\newblock Convolutional neural networks on graphs with fast localized spectral
  filtering.
\newblock In \emph{Advances in Neural Information Processing Systems}, pp.\
  3844--3852, 2016.

\bibitem[Duvenaud et~al.(2015)Duvenaud, Maclaurin, Iparraguirre, Bombarell,
  Hirzel, Aspuru-Guzik, and Adams]{duvenaud2015convolutional}
Duvenaud, D.~K., Maclaurin, D., Iparraguirre, J., Bombarell, R., Hirzel, T.,
  Aspuru-Guzik, A., and Adams, R.~P.
\newblock Convolutional networks on graphs for learning molecular fingerprints.
\newblock In \emph{Advances in neural information processing systems}, pp.\
  2224--2232, 2015.

\bibitem[Gao \& Ji(2019)Gao and Ji]{gao2019graph}
Gao, H. and Ji, S.
\newblock Graph u-net.
\newblock \emph{International Conference on Machine Learning}, 2019.

\bibitem[Gilmer et~al.(2017)Gilmer, Schoenholz, Riley, Vinyals, and
  Dahl]{gilmer2017neural}
Gilmer, J., Schoenholz, S.~S., Riley, P.~F., Vinyals, O., and Dahl, G.~E.
\newblock Neural message passing for quantum chemistry.
\newblock \emph{arXiv preprint arXiv:1704.01212}, 2017.

\bibitem[G{\'o}mez-Bombarelli et~al.(2018)G{\'o}mez-Bombarelli, Wei, Duvenaud,
  Hern{\'a}ndez-Lobato, S{\'a}nchez-Lengeling, Sheberla, Aguilera-Iparraguirre,
  Hirzel, Adams, and Aspuru-Guzik]{gomez2016automatic}
G{\'o}mez-Bombarelli, R., Wei, J.~N., Duvenaud, D., Hern{\'a}ndez-Lobato,
  J.~M., S{\'a}nchez-Lengeling, B., Sheberla, D., Aguilera-Iparraguirre, J.,
  Hirzel, T.~D., Adams, R.~P., and Aspuru-Guzik, A.
\newblock Automatic chemical design using a data-driven continuous
  representation of molecules.
\newblock \emph{ACS Central Science}, 2018.
\newblock \doi{10.1021/acscentsci.7b00572}.

\bibitem[Guimaraes et~al.(2017)Guimaraes, Sanchez-Lengeling, Farias, and
  Aspuru-Guzik]{guimaraes2017objective}
Guimaraes, G.~L., Sanchez-Lengeling, B., Farias, P. L.~C., and Aspuru-Guzik, A.
\newblock Objective-reinforced generative adversarial networks (organ) for
  sequence generation models.
\newblock \emph{arXiv preprint arXiv:1705.10843}, 2017.

\bibitem[Hamilton et~al.(2017)Hamilton, Ying, and
  Leskovec]{hamilton2017inductive}
Hamilton, W.~L., Ying, R., and Leskovec, J.
\newblock Inductive representation learning on large graphs.
\newblock \emph{arXiv preprint arXiv:1706.02216}, 2017.

\bibitem[Jin et~al.(2018)Jin, Barzilay, and Jaakkola]{jin2018junction}
Jin, W., Barzilay, R., and Jaakkola, T.
\newblock Junction tree variational autoencoder for molecular graph generation.
\newblock \emph{International Conference on Machine Learning}, 2018.

\bibitem[Jin et~al.(2019)Jin, Yang, Barzilay, and Jaakkola]{jin2018learning}
Jin, W., Yang, K., Barzilay, R., and Jaakkola, T.
\newblock Learning multimodal graph-to-graph translation for molecular
  optimization.
\newblock \emph{International Conference on Learning Representations}, 2019.

\bibitem[Kang \& Cho(2018)Kang and Cho]{kang2018conditional}
Kang, S. and Cho, K.
\newblock Conditional molecular design with deep generative models.
\newblock \emph{Journal of chemical information and modeling}, 59\penalty0
  (1):\penalty0 43--52, 2018.

\bibitem[Kearnes et~al.(2016)Kearnes, McCloskey, Berndl, Pande, and
  Riley]{kearnes2016molecular}
Kearnes, S., McCloskey, K., Berndl, M., Pande, V., and Riley, P.
\newblock Molecular graph convolutions: moving beyond fingerprints.
\newblock \emph{Journal of computer-aided molecular design}, 30\penalty0
  (8):\penalty0 595--608, 2016.

\bibitem[Kingma \& Welling(2013)Kingma and Welling]{kingma2013auto}
Kingma, D.~P. and Welling, M.
\newblock Auto-encoding variational bayes.
\newblock \emph{arXiv preprint arXiv:1312.6114}, 2013.

\bibitem[Kipf \& Welling(2017)Kipf and Welling]{kipf2016semi}
Kipf, T.~N. and Welling, M.
\newblock Semi-supervised classification with graph convolutional networks.
\newblock \emph{International Conference on Learning Representations}, 2017.

\bibitem[Kusner et~al.(2017)Kusner, Paige, and
  Hern{\'a}ndez-Lobato]{kusner2017grammar}
Kusner, M.~J., Paige, B., and Hern{\'a}ndez-Lobato, J.~M.
\newblock Grammar variational autoencoder.
\newblock \emph{arXiv preprint arXiv:1703.01925}, 2017.

\bibitem[Lei et~al.(2017)Lei, Jin, Barzilay, and Jaakkola]{lei2017deriving}
Lei, T., Jin, W., Barzilay, R., and Jaakkola, T.
\newblock Deriving neural architectures from sequence and graph kernels.
\newblock \emph{International Conference on Machine Learning}, 2017.

\bibitem[Li et~al.(2015)Li, Tarlow, Brockschmidt, and Zemel]{li2015gated}
Li, Y., Tarlow, D., Brockschmidt, M., and Zemel, R.
\newblock Gated graph sequence neural networks.
\newblock \emph{arXiv preprint arXiv:1511.05493}, 2015.

\bibitem[Li et~al.(2018)Li, Vinyals, Dyer, Pascanu, and
  Battaglia]{li2018learning}
Li, Y., Vinyals, O., Dyer, C., Pascanu, R., and Battaglia, P.
\newblock Learning deep generative models of graphs.
\newblock \emph{arXiv preprint arXiv:1803.03324}, 2018.

\bibitem[Liao et~al.(2019)Liao, Li, Song, Wang, Hamilton, Duvenaud, Urtasun,
  and Zemel]{liao2019efficient}
Liao, R., Li, Y., Song, Y., Wang, S., Hamilton, W., Duvenaud, D.~K., Urtasun,
  R., and Zemel, R.
\newblock Efficient graph generation with graph recurrent attention networks.
\newblock In \emph{Advances in Neural Information Processing Systems}, pp.\
  4257--4267, 2019.

\bibitem[Liu et~al.(2018)Liu, Allamanis, Brockschmidt, and
  Gaunt]{liu2018constrained}
Liu, Q., Allamanis, M., Brockschmidt, M., and Gaunt, A.~L.
\newblock Constrained graph variational autoencoders for molecule design.
\newblock \emph{Neural Information Processing Systems}, 2018.

\bibitem[Luong et~al.(2015)Luong, Pham, and Manning]{luong2015effective}
Luong, M.-T., Pham, H., and Manning, C.~D.
\newblock Effective approaches to attention-based neural machine translation.
\newblock \emph{arXiv preprint arXiv:1508.04025}, 2015.

\bibitem[Ma et~al.(2018)Ma, Chen, and Xiao]{ma2018constrained}
Ma, T., Chen, J., and Xiao, C.
\newblock Constrained generation of semantically valid graphs via regularizing
  variational autoencoders.
\newblock In \emph{Advances in Neural Information Processing Systems}, pp.\
  7113--7124, 2018.

\bibitem[Niepert et~al.(2016)Niepert, Ahmed, and Kutzkov]{niepert2016learning}
Niepert, M., Ahmed, M., and Kutzkov, K.
\newblock Learning convolutional neural networks for graphs.
\newblock In \emph{International Conference on Machine Learning}, pp.\
  2014--2023, 2016.

\bibitem[Olivecrona et~al.(2017)Olivecrona, Blaschke, Engkvist, and
  Chen]{olivecrona2017molecular}
Olivecrona, M., Blaschke, T., Engkvist, O., and Chen, H.
\newblock Molecular de-novo design through deep reinforcement learning.
\newblock \emph{Journal of cheminformatics}, 9\penalty0 (1):\penalty0 48, 2017.

\bibitem[Polykovskiy et~al.(2018)Polykovskiy, Zhebrak, Sanchez-Lengeling,
  Golovanov, Tatanov, Belyaev, Kurbanov, Artamonov, Aladinskiy, Veselov,
  Kadurin, Nikolenko, Aspuru-Guzik, and Zhavoronkov]{polykovskiy2018molecular}
Polykovskiy, D., Zhebrak, A., Sanchez-Lengeling, B., Golovanov, S., Tatanov,
  O., Belyaev, S., Kurbanov, R., Artamonov, A., Aladinskiy, V., Veselov, M.,
  Kadurin, A., Nikolenko, S., Aspuru-Guzik, A., and Zhavoronkov, A.
\newblock {M}olecular {S}ets ({MOSES}): {A} {B}enchmarking {P}latform for
  {M}olecular {G}eneration {M}odels.
\newblock \emph{arXiv preprint arXiv:1811.12823}, 2018.

\bibitem[Popova et~al.(2018)Popova, Isayev, and Tropsha]{popova2018deep}
Popova, M., Isayev, O., and Tropsha, A.
\newblock Deep reinforcement learning for de novo drug design.
\newblock \emph{Science advances}, 4\penalty0 (7):\penalty0 eaap7885, 2018.

\bibitem[Rogers \& Hahn(2010)Rogers and Hahn]{rogers2010extended}
Rogers, D. and Hahn, M.
\newblock Extended-connectivity fingerprints.
\newblock \emph{Journal of chemical information and modeling}, 50\penalty0
  (5):\penalty0 742--754, 2010.

\bibitem[Samanta et~al.(2018)Samanta, De, Jana, Chattaraj, Ganguly, and
  Gomez-Rodriguez]{samanta2018nevae}
Samanta, B., De, A., Jana, G., Chattaraj, P.~K., Ganguly, N., and
  Gomez-Rodriguez, M.
\newblock Nevae: A deep generative model for molecular graphs.
\newblock \emph{arXiv preprint arXiv:1802.05283}, 2018.

\bibitem[Scarselli et~al.(2009)Scarselli, Gori, Tsoi, Hagenbuchner, and
  Monfardini]{scarselli2009graph}
Scarselli, F., Gori, M., Tsoi, A.~C., Hagenbuchner, M., and Monfardini, G.
\newblock The graph neural network model.
\newblock \emph{IEEE Transactions on Neural Networks}, 20\penalty0
  (1):\penalty0 61--80, 2009.

\bibitem[Sch{\"u}tt et~al.(2017)Sch{\"u}tt, Kindermans, Felix, Chmiela,
  Tkatchenko, and M{\"u}ller]{schutt2017schnet}
Sch{\"u}tt, K., Kindermans, P.-J., Felix, H. E.~S., Chmiela, S., Tkatchenko,
  A., and M{\"u}ller, K.-R.
\newblock Schnet: A continuous-filter convolutional neural network for modeling
  quantum interactions.
\newblock In \emph{Advances in Neural Information Processing Systems}, pp.\
  992--1002, 2017.

\bibitem[Seff et~al.(2019)Seff, Zhou, Damani, Doyle, and
  Adams]{seff2019discrete}
Seff, A., Zhou, W., Damani, F., Doyle, A., and Adams, R.~P.
\newblock Discrete object generation with reversible inductive construction.
\newblock In \emph{Advances in Neural Information Processing Systems}, pp.\
  10353--10363, 2019.

\bibitem[Segler et~al.(2017)Segler, Kogej, Tyrchan, and
  Waller]{segler2017generating}
Segler, M.~H., Kogej, T., Tyrchan, C., and Waller, M.~P.
\newblock Generating focussed molecule libraries for drug discovery with
  recurrent neural networks.
\newblock \emph{arXiv preprint arXiv:1701.01329}, 2017.

\bibitem[Simonovsky \& Komodakis(2018)Simonovsky and
  Komodakis]{simonovsky2018graphvae}
Simonovsky, M. and Komodakis, N.
\newblock Graphvae: Towards generation of small graphs using variational
  autoencoders.
\newblock \emph{arXiv preprint arXiv:1802.03480}, 2018.

\bibitem[St.~John et~al.(2019)St.~John, Phillips, Kemper, Wilson, Guan,
  Crowley, Nimlos, and Larsen]{st2019message}
St.~John, P.~C., Phillips, C., Kemper, T.~W., Wilson, A.~N., Guan, Y., Crowley,
  M.~F., Nimlos, M.~R., and Larsen, R.~E.
\newblock Message-passing neural networks for high-throughput polymer
  screening.
\newblock \emph{The Journal of chemical physics}, 150\penalty0 (23):\penalty0
  234111, 2019.

\bibitem[Velickovic et~al.(2017)Velickovic, Cucurull, Casanova, Romero, Lio,
  and Bengio]{velickovic2017graph}
Velickovic, P., Cucurull, G., Casanova, A., Romero, A., Lio, P., and Bengio, Y.
\newblock Graph attention networks.
\newblock \emph{arXiv preprint arXiv:1710.10903}, 2017.

\bibitem[Weininger(1988)]{weininger1988smiles}
Weininger, D.
\newblock Smiles, a chemical language and information system. 1. introduction
  to methodology and encoding rules.
\newblock \emph{Journal of chemical information and computer sciences},
  28\penalty0 (1):\penalty0 31--36, 1988.

\bibitem[Xu et~al.(2018)Xu, Hu, Leskovec, and Jegelka]{xu2018powerful}
Xu, K., Hu, W., Leskovec, J., and Jegelka, S.
\newblock How powerful are graph neural networks?
\newblock \emph{arXiv preprint arXiv:1810.00826}, 2018.

\bibitem[Ying et~al.(2018)Ying, You, Morris, Ren, Hamilton, and
  Leskovec]{ying2018hierarchical}
Ying, Z., You, J., Morris, C., Ren, X., Hamilton, W., and Leskovec, J.
\newblock Hierarchical graph representation learning with differentiable
  pooling.
\newblock In \emph{Advances in Neural Information Processing Systems}, pp.\
  4800--4810, 2018.

\bibitem[You et~al.(2018{\natexlab{a}})You, Liu, Ying, Pande, and
  Leskovec]{you2018graph}
You, J., Liu, B., Ying, R., Pande, V., and Leskovec, J.
\newblock Graph convolutional policy network for goal-directed molecular graph
  generation.
\newblock \emph{arXiv preprint arXiv:1806.02473}, 2018{\natexlab{a}}.

\bibitem[You et~al.(2018{\natexlab{b}})You, Ying, Ren, Hamilton, and
  Leskovec]{you2018graphrnn}
You, J., Ying, R., Ren, X., Hamilton, W.~L., and Leskovec, J.
\newblock Graphrnn: A deep generative model for graphs.
\newblock \emph{arXiv preprint arXiv:1802.08773}, 2018{\natexlab{b}}.

\bibitem[Zhou et~al.(2018)Zhou, Kearnes, Li, Zare, and
  Riley]{zhou2018optimization}
Zhou, Z., Kearnes, S., Li, L., Zare, R.~N., and Riley, P.
\newblock Optimization of molecules via deep reinforcement learning.
\newblock \emph{arXiv preprint arXiv:1810.08678}, 2018.

\bibitem[Zhu et~al.(2017)Zhu, Zhang, Pathak, Darrell, Efros, Wang, and
  Shechtman]{zhu2017toward}
Zhu, J.-Y., Zhang, R., Pathak, D., Darrell, T., Efros, A.~A., Wang, O., and
  Shechtman, E.
\newblock Toward multimodal image-to-image translation.
\newblock In \emph{Advances in Neural Information Processing Systems}, pp.\
  465--476, 2017.

\end{thebibliography}
\bibliographystyle{icml2020}

\newpage
\appendix
\onecolumn

\section{Motif Construction}

\begin{figure}[h]
    \centering
    \includegraphics[width=\textwidth]{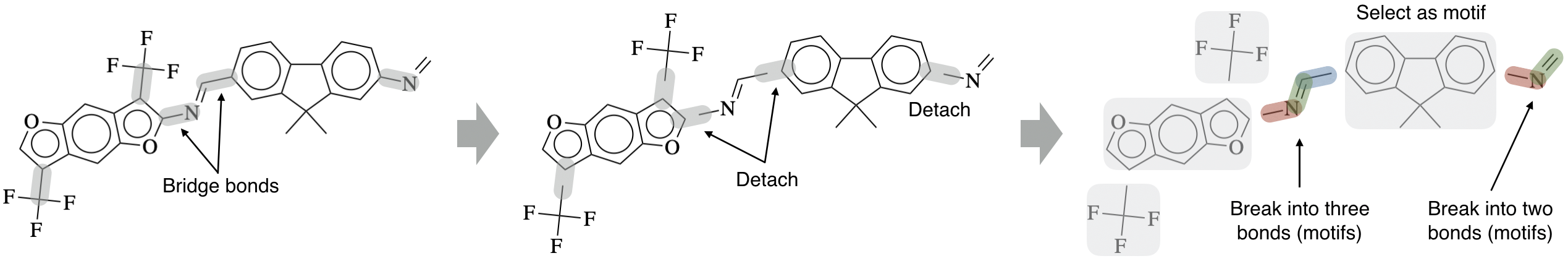}
    \caption{Illustration of motif extraction procedure.}
    \vspace{-3pt}
    \label{fig:extract}
    \vspace{-3pt}
\end{figure}

To extract motifs, we decompose a molecule $\graph=(\gV,\gE)$ into disconnected fragments by breaking all the bridge bonds that will not violate chemical validity. Our motif extraction consists of three steps (see Figure~\ref{fig:extract}):
\begin{enumerate}[leftmargin=*,topsep=0pt,itemsep=0pt]
    \item Find all the \emph{bridge} bonds $(u,v) \in \gE$, where both $u$ and $v$ have degree $\Delta_u, \Delta_v \geq 2$ and either $u$ or $v$ is part of a ring.
    \item Detach all the bridge bonds from its neighbors. Now the graph $\graph$ becomes a set of disconnected subgraphs $\graph_1, \cdots, \graph_N$. 
    \item Select $\graph_i$ as motif in $\graph$ if its occurrence in the training set is more than $\Delta=100$. If $\graph_i$ is not selected as motif, further decompose it into rings and bonds and put them into the motif vocabulary $V_\gS$.
\end{enumerate}

\section{Network Architecture}

\textbf{MPN Architecture } Our message passing network $\MPN_{\psi}\left(\gH, \set{\vx_u}, \set{\vx_{uv}} \right)$ is a slight modification from the MPN architecture used in \citet{dai2016discriminative,jin2018learning}. Let $N(v)$ be the neighbors of node $v$, $\vx_v$ the node feature of $v$ and $\vx_{uv}$ be the feature of edge $(u,v)$. During encoding, each edge $(u,v)$ is associated with two messages $\vnu_{uv}$ and $\vnu_{vu}$, representing the message from $u$ to $v$ and vice versa. The messages are updated by an LSTM cell with parameters  $\psi=\set{\mW_\psi^z,\mW_\psi^o,\mW_\psi^r,\mW_\psi}$ defined as follows: 
\begin{algorithm}
\caption{LSTM Message Passing}
\begin{algorithmic}
\FUNCTION{$\LSTM_\psi \left(\vx_u, \vx_{uv}, \set{\vnu_{wu}^{(t)}, \vc_{wu}^{(t)}}_{w \in N(u)\backslash v}\right)$  }
\STATE \vspace{-10pt}\begin{flalign}
\begin{aligned}
\vi_{uv} &= \sigma\left(\mW_\psi^z \left[\vx_u, \vx_{uv}, \sum\nolimits_w \vnu_{wu}^{(t)} \right] + \vb^z\right) \qquad \qquad \vo_{uv} = \sigma\left(\mW_\psi^o \left[\vx_u, \vx_{uv}, \sum\nolimits_w \vnu_{wu}^{(t)} \right] + \vb^o\right) \\
\vf_{wu} &= \sigma\left(\mW_\psi^r \left[\vx_u, \vx_{uv}, \vnu_{wu}^{(t)} \right] + \vb^r\right) \qquad \qquad \quad\; \tilde{\vc}_{uv}^{(t+1)} = \tanh \left(\mW_\psi \left[\vx_u, \vx_{uv}, \sum\nolimits_w \vnu_{wu}^{(t)} \right] + \vb \right) \\
\vc_{uv}^{(t+1)} &= \vi_{uv} \odot \tilde{\vc}_{uv}^{(t+1)} + \sum\nolimits_w \vf_{wu} \odot \vc_{wu}^{(t)}  \qquad \qquad \;\; \vnu_{uv}^{(t+1)} = \vo_{uv} \odot \tanh\left( \vc_{uv}^{(t+1)}\right)
\end{aligned} && \nonumber
\end{flalign}
\vspace{-5pt}
\STATE \textbf{Return } $\vnu_{uv}^{(t+1)},\vc_{uv}^{(t+1)}$
\vspace{3pt}
\ENDFUNCTION
\STATE
\FUNCTION{$\MPN_\psi \left(\gH, \set{\vx_v}, \set{\vx_{uv}}\right)$}
\STATE Initialize messages: $\vnu_{uv}^{0} = \mathbf{0}, \vc_{uv}^{0} = \mathbf{0}$
\FOR{$t=0$ \textbf{to} $T-1$}
\STATE Compute messages $\vnu_{uv}^{(t+1)}, \vc_{uv}^{(t+1)} = \LSTM_{\psi}\left(\vx_u, \vx_{uv}, \set{ \vnu_{wu}^{(t)}, \vc_{wu}^{(t)}}_{w \in N(u) \backslash v}\right)$ for all edges $(u,v)\in\gH$.
\ENDFOR
\STATE \textbf{Return } node representations $\vh_v = \MLP\left(\vx_v, \sum_{u \in N(v)} \vnu_{uv}^{(T)}\right)$
\vspace{3pt}
\ENDFUNCTION
\end{algorithmic}
\end{algorithm}

%\textbf{Attention Layer } Our attention layer is a bilinear attention function with %parameter $\theta=\set{\mA_\theta}$:
%\begin{equation}
%    \attention_{\theta}(\vv, \set{\vh_i}) = \sum_i \alpha_i \vh_i  \qquad \alpha_i = \frac{\exp(\vv^T \mA_\theta \vh_i)}{\sum_j \exp(\vv^T \mA_\theta \vh_j)}
%\end{equation}

\begin{figure}[t]
    \centering
    \includegraphics[width=\textwidth]{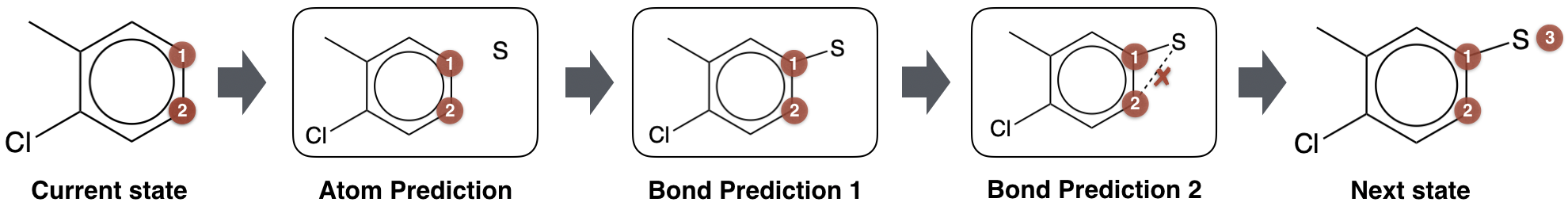}
    \caption{Illustration of AtomG2G decoding process. Atoms marked with red circles are frontier nodes in the queue $\gQ$. In each step, the model picks the first node $v_t$ from $\gQ$ and predict whether there will be new atoms attached to $v_t$. If so, it predicts the atom type of new node $u_t$ (atom prediction). Then the model predicts the bond type between $u_t$ and other nodes in $\gQ$ sequentially for $|\gQ|$ steps (bond prediction, $|\gQ|=2$). Finally, it adds the new atom to the queue $\gQ$.}
    \label{fig:atomg2g}
\end{figure}

\textbf{AtomG2G Architecture } AtomG2G is an atom-based translation method that is directly comparable to HierG2G. Here molecules are represented solely as molecular graphs rather than a hierarchical graph with motifs. The encoder of AtomG2G uses the same LSTM MPN to encode molecular graph. This gives us a set of atom vectors $\vc_X^\graph$ representing molecule $X$ only at the atom level.
The decoder of AtomG2G is illustrated in Figure~\ref{fig:atomg2g}. Following \citet{you2018graphrnn,liu2018constrained}, the model generates molecule $\graph$ atom by atom following their breadth-first order. During generation, it maintains a FIFO queue $\gQ$ that contains the frontier nodes in the graph (i.e., nodes who still have neighbors to be generated). 
Let $v_t$ be the first node in $\gQ$ and $\graph_t$ be the current graph at step $t$. In each step, the model makes three predictions to expand the graph $\graph_t$:
\begin{enumerate}[leftmargin=*,topsep=0pt,itemsep=0pt]
\item It predicts whether there will be new atoms attached to $v_t$. If not, the model discards $v$ and move on to the next node in $\gQ$. The generation stops if $\gQ$ is empty.
\item Otherwise, it creates a new atom $u_t$ and predicts its atom type.
\item Lastly, it predicts the bond type between $u_t$ and other frontier nodes in $\gQ$ autoregressively to fully capture edge dependencies. Since nodes are generated in breadth-first order, there will be no edges between $u_t$ and nodes outside of $\gQ$.
\end{enumerate}

To make those predictions, we use the same LSTM MPN to encode the current graph $\graph_t$. Let $\vh_{v_t}$ be the atom representation of $v_t$. We represent $\graph_t$ as the sum of all its atom vectors $\vh_{\graph_t} = \sum_{v \in \graph_t} \vh_v$. In the first step, we model the probability of expanding a new node from $v_t$ as:
\begin{equation}
    \vp_t = \sigmoid(\MLP(\vh_{v_t}, \vh_{\graph_t}, \valpha_t^d)) \qquad \valpha_t^d = \attention_d\left( [\vh_{v_t},\vh_{\graph_t}], \vc_X^\graph \right)
\end{equation}
In the second step, the atom type of the new node $u_t$ is predicted using another MLP:
\begin{equation}
    \vq_t = \softmax(\MLP(\vh_{v_t}, \vh_{\graph_t}, \valpha_t^s)) \qquad \valpha_t^s = \attention_s\left( [\vh_{v_t},\vh_{\graph_t}], \vc_X^\graph \right)
\end{equation}
In the last step, we predict the bonds between $u_t$ and nodes in $\gQ={a_1,\cdots,a_n}$ sequentially starting with $a_1=v_t$. Specifically, for each atom pair $(u_t,a_k)$, we predict their bond type (single, double, triple or none) as the following:
\begin{equation}
    \vb_{u_t, a_k} = \softmax( \MLP(\vh_{\graph_t}, \vh_{u_t}^k, \vh_{a_k}, \valpha_t^b) ) \qquad
    \valpha_t^b = \attention_b \left( [\vh_{\graph_t}, \vh_{u_t}^k, \vh_{a_k}], \vc_X^\graph \right)
\end{equation}
where $\vh_{a_k}$ is the atom representation of node $a_k$ and $\vh_{u_t}^k$ is the representation of node $u_t$ at the $k^{\textrm{th}}$ bond prediction. Let $N_k(u_t)$ be node $u_t$'s current neighbor predicted in the first $k$ steps. $\vh_{u_t}^k$ is computed as follows to reflect its local graph structure after $k^{\textrm{th}}$ bond prediction:
\begin{equation}
    \vh_{u_t}^k = \MLP\left(\vx_{u_t}, \sum\nolimits_{w \in N_k(u_t)} \vnu_{w,u_t} \right) \qquad \vnu_{w,u_t} = \MLP(\vh_w, \vx_{w,u_t})
\end{equation}
where $\vx_{u_t}$ is the atom feature of $u_t$ (i.e., predicted atom type) and $\vx_{w,u_t}$ is the bond feature between $w$ and $u_t$ (i.e., predicted bond type). Intuitively, this can be viewed as running one-step message passing at each bond prediction step (i.e., passing the message $\vnu_{w,u_t}$ from $w$ to $u_t$).
AtomG2G is trained under the same variational objective as HierG2G, with the latent code $\vz$ sampled from the posterior $Q(\vz | X, Y)=\gN(\vmu_{X,Y}, \vsigma_{X,Y})$ and $[\vmu_{X,Y}, \vsigma_{X,Y}] = \MLP(\sum \vc_Y^\graph - \sum \vc_X^\graph)$.

\section{Experimental Details}
\begin{table}[t]
    \centering
    \begin{tabular}{lccccc}
        \hline
        & Polymer & logP ($\delta=0.6$) & logP ($\delta=0.4$) & QED & DRD2 \Tstrut\Bstrut \\
        \hline
        Training set size & 76K & 75K & 99K & 88K & 34K \Tstrut\Bstrut \\
        Test set size & 5000 & 800 & 800 & 800 & 1000 \Tstrut\Bstrut \\
        Motif vocabulary size $|\gS|$ & 436 & 478 & 462 & 307 & 307 \Tstrut\Bstrut \\
        Attachment vocabulary size (avg.) $|\gA(\gS_t)|$ & 5.24 & 3.68 & 3.50 & 3.62 & 3.30 \Tstrut\Bstrut \\ \hline
    \end{tabular}
    \caption{Training set size and motif vocabulary size for each dataset.}
    \label{tab:data}
\end{table}

\begin{figure}[t]
    \centering
    \includegraphics[width=\textwidth]{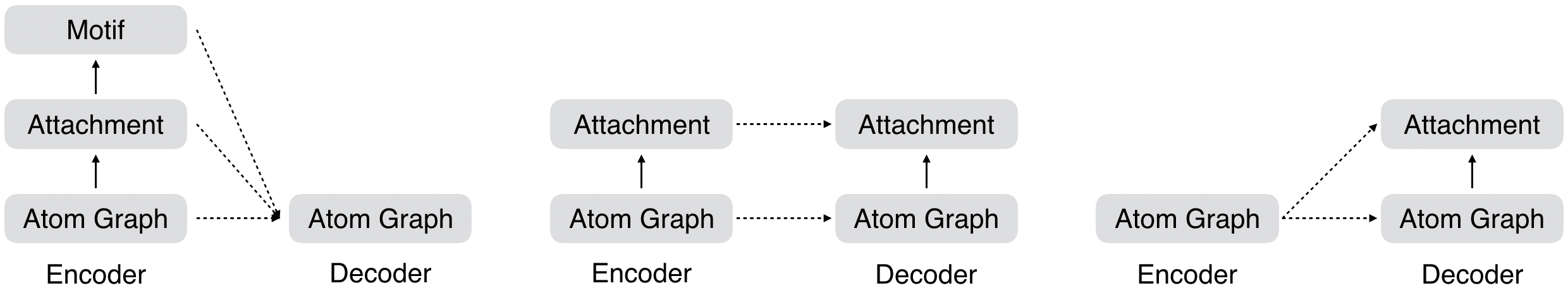}
    \caption{Ablation studies in graph translation tasks. \textbf{Left}: Atom-based decoder; \textbf{Middle}: Two-layer encoder; \textbf{Right}: One-layer encoder.}
    \label{fig:ablation}
\end{figure}

\begin{figure}[th!]
    \centering
    \includegraphics[width=0.75\textwidth]{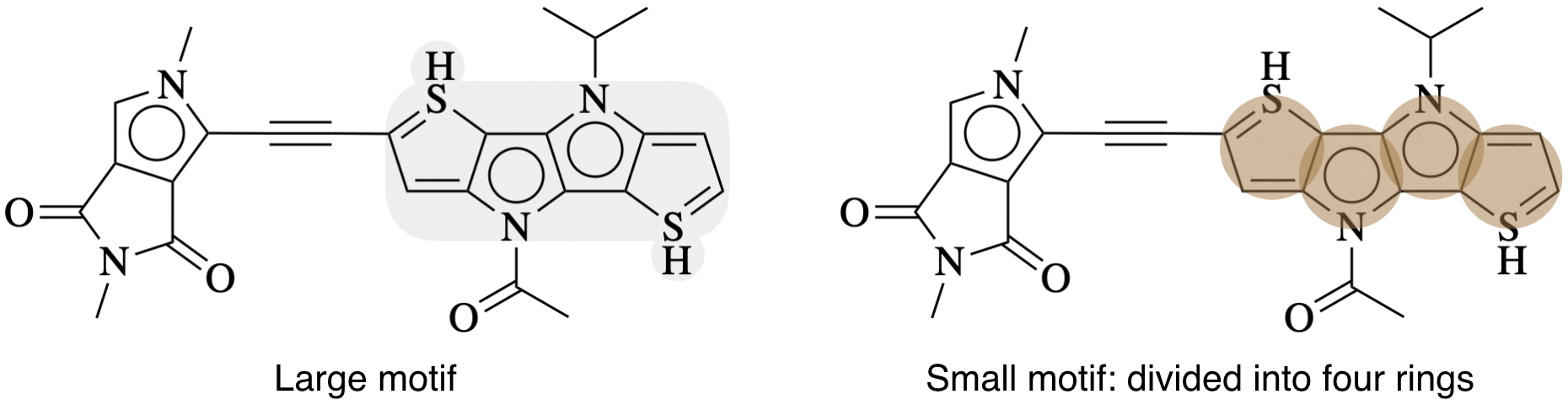}
    \caption{Ablation study of HierVAE (coined as $\mathrm{small \; motif}$), where motifs are restricted to single rings and bonds.}
    \label{fig:small-motif}
\end{figure}

\subsection{Polymer Generation}
\textbf{Data } The polymer dataset~\citep{st2019message} is downloaded from \url{https://cscdata.nrel.gov/#/datasets/ad5d2c9a-af0a-4d72-b943-1e433d5750d6}. The dataset and motif vocabulary sizes are listed in Table~\ref{tab:data}.

\textbf{Hyperparameters } For HierVAE, we set the hidden layer dimension to be 400 and latent code dimension $|\vz|=20$. For CG-VAE and JT-VAE, we used their official implementations for our experiments with their suggested hyperparameters. We set the KL regularization weight $\lambda_{\textrm{KL}}=0.1$ for all models. Each model has around 5M parameters.

\begin{figure}[t]
    \centering
    \includegraphics[width=0.97\textwidth]{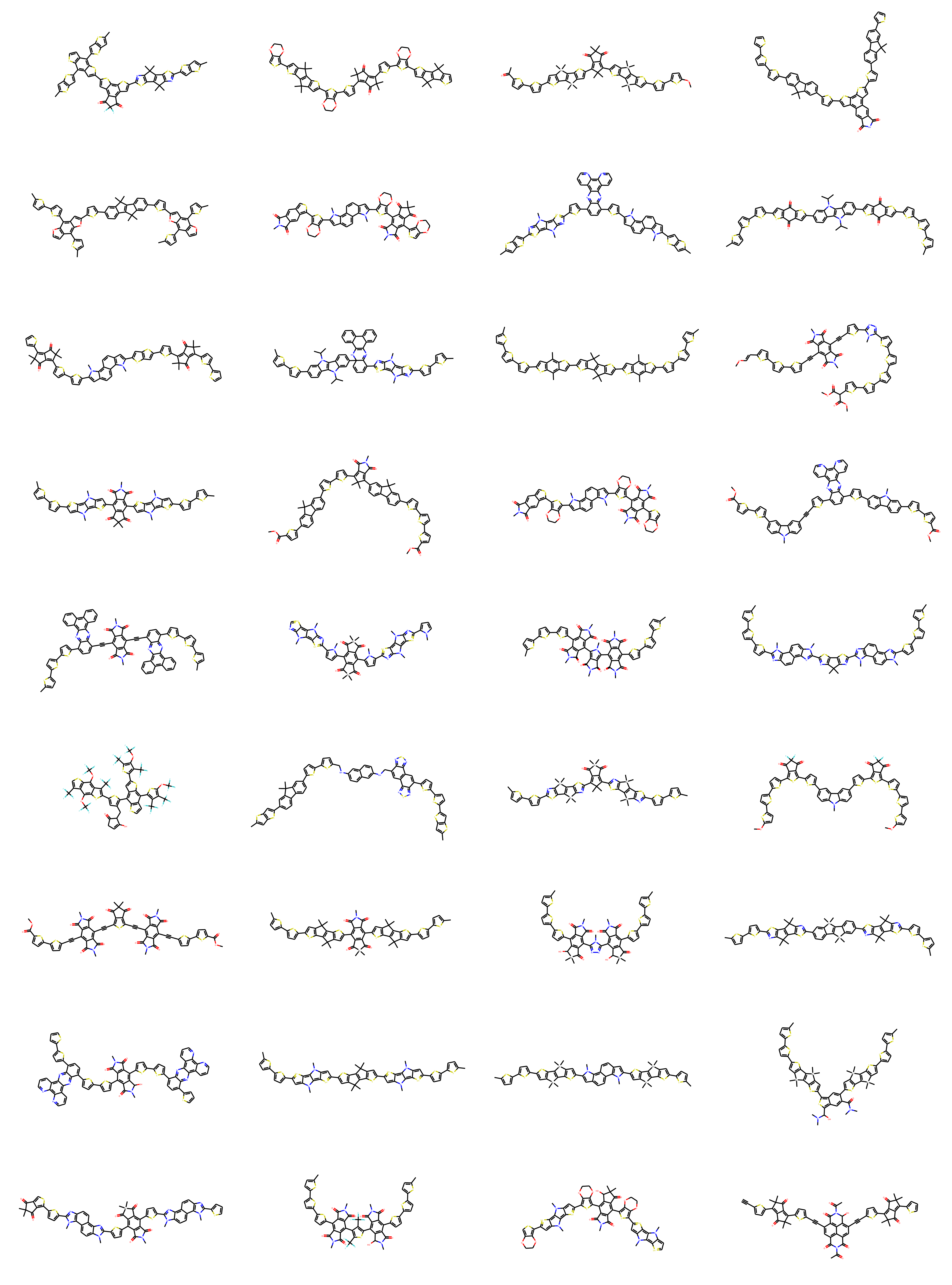}
    \vspace{-10pt}
    \caption{Sampled polymers from HierVAE.}
    \label{fig:samples}
    \vspace{-5pt}
\end{figure}

\textbf{Metrics and Samples} Our metrics are computed using the implementation from \citet{polykovskiy2018molecular} (\url{https://github.com/molecularsets/moses}). Samples from our models are shown in Figure~\ref{fig:samples}.

\textbf{Ablation Study } Our $\mathrm{small \; motif}$ baseline builds on the same hierarchical architecture as our model, but it generates molecules based on small motifs restricted to single rings or bonds (see Figure~\ref{fig:small-motif}).

\subsection{Graph-to-Graph Translation}
\textbf{Data } The graph translation datasets are directly downloaded from the link provided in \citet{jin2018learning}. The dataset and motif vocabulary size for each dataset is listed in Table~\ref{tab:data}.

\textbf{Hyperparameters } For HierG2G, we set the hidden layer dimension to be 270 and the embedding layer dimension 200. We set the latent code dimension $|\vz|=8$ and KL regularization weight $\lambda_{\textrm{KL}}=0.3$. We run $T=20$ iterations of message passing in each layer of the encoder. 
For AtomG2G, we set the hidden layer and embedding layer dimension to be 400 so that both models have roughly the same number of parameters. We also set $\lambda_{\textrm{KL}}=0.3$ and number of message passing iterations to be $T=20$. We train both models with Adam optimizer with default parameters.

\textbf{Ablation Study } Our ablation studies are illustrated in Figure~\ref{fig:ablation}. In our first experiment, we changed our decoder to the atom-based decoder of AtomG2G. As the encoder is still hierarchical, we modified the input of the decoder attention to include both atom and motif vectors. We set the hidden layer and embedding layer dimension to be 300 to match the original model size. 
Our next two experiments reduces the number of hierarchies in both our encoder and decoder MPN. In the two-layer model, molecules are represented by $\vc_X = \vc_X^\graph \cup \vc_X^\gA$. We make motif predictions based on hidden vector $\vh_{\gA_k}$ instead of $\vh_{\gS_k}$ because the motif layer is removed. 
In the one-layer model, molecules are represented by $\vc_X = \vc_X^\graph$ and we make motif predictions based on atom vectors $\sum_{v\in \gS_k} \vh_v$. The hidden layer dimension is adjusted accordingly to match the original model size.

\end{document}